\documentclass[sigconf,9pt]{acmart}

\usepackage{balance} 
\usepackage{booktabs} % For formal tables
\usepackage{multirow}
\usepackage{subcaption}
\usepackage{algorithm}
\usepackage[english]{babel}
\usepackage{lipsum}
\usepackage{mathtools}
\usepackage{cuted}
\usepackage{gensymb}
\usepackage{amsmath} 
\usepackage{booktabs}
\usepackage{algpseudocode}
\usepackage{color}

\usepackage{todonotes}

\begin{document}

 \title{Estimating Buildings' Parameters over Time Including Prior Knowledge }

%\title{Estimating Buildings' Parameters over Time Including Prior Knowledge}
    \author{Nilavra Pathak}
    \affiliation{%
    \department{Information Systems}
        \institution{University of Maryland Baltimore County}
        }
    \email{nilavra1@umbc.edu}
       \author{James Foulds}
    \affiliation{%
           \department{Information Systems}
        \institution{University of Maryland Baltimore County}
           }
    \email{jfoulds@umbc.edu}
       \author{Nirmalya Roy}
    \affiliation{%
         \department{Information Systems}
        \institution{University of Maryland Baltimore County}
            }
    \email{nroy@umbc.edu}
       \author{Nilanjan Banerjee}
    \affiliation{%
            \department{Computer Science and Electrical Engineering}
        \institution{University of Maryland Baltimore County}
          }
    \email{nilanb@umbc.edu}
       \author{Ryan Robucci}
    \affiliation{%
                \department{Computer Science and Electrical Engineering}
        \institution{University of Maryland Baltimore County}
          }
    \email{robucci@umbc.edu}

\begin{abstract}
Modeling buildings' heat dynamics is a complex process which depends on various factors including weather, building thermal capacity, insulation preservation, and residents' behavior. Gray-box models offer an explanation of those dynamics, as expressed in a few parameters specific to built environments. These parameters can provide compelling insights into the characteristics of building artifacts and have various applications such as forecasting HVAC usage, indoor temperature control monitoring of built environments, and more. In this paper, we present a systematic study of Bayesian approaches to modeling buildings' parameters, and hence their thermal characteristics. We build a Bayesian state-space model that can adapt and incorporate buildings' thermal equations and postulate a generalized solution that can easily adapt prior knowledge regarding the parameters. We then show that a faster approximate approach using Variational Inference for parameter estimation can posit similar parameters' quantification as that of a more time-consuming Markov Chain Monte Carlo (MCMC) approach. We perform extensive evaluations on two datasets to understand the generative process and attest that the Bayesian approach is more interpretable. We further study the effects of prior selection on the model parameters and transfer learning, where we learn parameters from one season and reuse them to fit the model in other seasons. We perform extensive evaluations on controlled and real data traces to enumerate buildings' parameters within a 95\% credible interval. 
\end{abstract}

 \pagenumbering{arabic}
 \maketitle

 \section{{Introduction}}
 Retrofitting an existing building often reduces its energy consumption and particularly the heating and cooling costs. To assess the effectiveness of the retrofit, auditors perform on-site tests to gauge the insulation and infiltration quality of a house. However, such tests are expensive and intrusive, and thus cannot be carried out continuously. The proliferation of smart thermostats %\jf{These are already here, right? I'd delete ``on the horizon''} 
 such as   NEST and Ecobee~\cite{KLEIMINGER2014493}, and their acceptance and deployment in home environments are opening up new research avenues. In the near term, we envision a self-adaptive and programmable thermostat, that can seamlessly receive environmental data from the indoor and outdoors, and residents' activities, to model the inherent thermal characteristics of the building. Such a dynamic and adaptive smart thermostat will provide an early assessment of the insulation and leakage, and thus help promote energy sensitive actions and maintain comfort levels. 
  
 \indent Physicists have studied methods for modeling buildings' thermal conditions by way of several measurable parameters~\cite{tarantola2005inverse,fabrizio2015methodologies}.  In these models, the  thermal dynamics of a building are represented by an RC-circuit, due to system equivalence,  which allows us to  derive a set of stochastic differential equations that describe the thermal patterns. The composite parameters \emph{resistance} (\textbf{R}) and \emph{capacitance} (\textbf{C}) of the circuit are analogous to the buildings' insulation (and to some extent the infiltration), and the thermal mass,  respectively. Building quality measurement uses standardized metrics such as R-value (or U-value) to measure insulation and $ACH_{50}$ to measure infiltration. The thermal mass of a house is the ability of a material to absorb and store heat energy. Optimization based techniques~\cite{Optimization,gouda2002building} are popularly used to estimate the parameters, where the objective is to reduce the error between observed and predicted values. However, most approaches do not simultaneously consider two key factors that are common in the real world:
    \begin{itemize}
   \item \textit{Stochasticity of the building parameters:} The optimization-based methods are effective for fitting a model to data, but cannot provide a margin of error on the estimation. This is important as stochasticity arises due to several unaccounted factors, including human activity and home appliance   usage, which cannot be directly quantified. 
\item \textit{Presence of prior knowledge:} It is common knowledge that older buildings have poor insulation. Studies~\cite{EIA} show that the  average  house size has increased with time, and that larger homes typically have better insulation quality. By incorporation of prior knowledge such as \textit{``How old is a building?''} or \textit{``How much square footage does it have?''}, these intuitions about a building's condition can potentially increase the accuracy of the estimated parameters.
\end{itemize}

  \indent  To address these concerns, the Bayesian approach is a natural and simple way to incorporate prior knowledge in the building thermal modeling framework which also approximates the factors influencing the model dynamics. It allows for comparisons among multiple candidate models instead of performing binary hypothesis tests on a single model. The Bayesian posterior distribution plays the role of Occam's razor, effectively penalizing an increase in model complexity, such as adding variables, while rewarding improvements in fit. However, the existing Bayesian approaches have a few notable limitations: \emph{(i)} Bayesian inference of the parameters is primarily performed with Markov Chain Monte Carlo (MCMC) algorithms~\cite{gordon1993novel, geweke2001bayesian} which take a long time to converge, and thus are not well suited for the case where model complexity and/or data size increase. \emph{(ii)} A majority of previous works applied uninformed normal priors and do not evaluate the effect of prior selection on model performance. As such the full benefit of a Bayesian statistical approach is not utilized. \emph{(iii)} Finally, most studies limit their scope to a single seasonal period, particularly in the winter when residents use the HVAC in heating mode, and do not study how the model parameters estimated in one season can be used to monitor the house longitudinally.
  \begin{figure}[h]
    \begin{subfigure}{.23\textwidth} \includegraphics[height=3.5cm,width=1\textwidth]{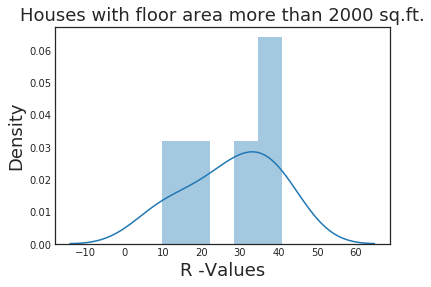}
    \end{subfigure}
    \begin{subfigure}{.23\textwidth}  \includegraphics[height=3.5cm,width=1\textwidth]{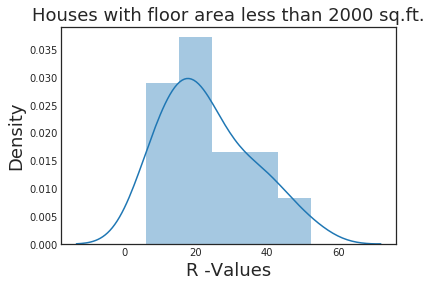}
    \end{subfigure}     
    \caption{Distribution of R-values of houses}
    \label{fig:hist}
    \end{figure}  
    
 \indent  To investigate these shortcomings and their resolution, in this paper we present a systematic study of Bayesian approaches to the modeling of buildings' thermal dynamics. We propose a generalized Bayesian State Space Model (BSSM) that can combine physics-based thermal models into a probabilistic framework. We further embed prior intuition and knowledge regarding buildings into the model based on subjective beliefs. For example, in Figure~\ref{fig:hist}, the probability densities of the R-values of homes built before the year 2000 differ depending on their size, in this case whether their area is less than 2000 square feet. We show how to incorporate such knowledge by effective prior selection. However, such priors are not conjugate to the likelihood and solutions cannot be computed analytically. We thus perform inference based on algorithms that do not depend on conjugacy, such as Automatic Differentiation Variational Inference, and show that the buildings' parameters can be estimated effectively with such an approximate approach. We analyze the effect of learning parameters from one season and use transfer learning to estimate the thermal dynamics in a different season when the HVAC is used in a different mode. We present two case studies on real data traces to show the effectiveness of the Bayesian approach, and the effects of prior selection and transfer learning across seasons.\\
\indent \textbf{Key Contributions:} Our innovations and results provide evidence that the Bayesian approach to modeling a building's thermal characteristics is valuable. The primary contributions of our work are as follows.  
\begin{itemize}
  \item \textit{Bayesian State Space Model:}  We propose a  Bayesian state-space model for estimating buildings' thermal parameters.  Unlike previous methods~\cite{bacher2011identifying,juhl2017grey, EstimationBuildsys,  KRISTENSEN2004225}  which use point estimates, our Bayesian model is capable of incorporating beliefs using non-conjugate priors, and managing uncertainty in the parameters.  We inferred the model parameters within a 95\% credible interval with a Mean Field Variational Approximation, and show that the estimates are as accurate as that of a more time-consuming MCMC approach. 
 %  \item \textit{Fast Bayesian inference using Variational Inference algorithms:} %\jf{I normally wouldn't capitalize Variational Inference, as it's not a proper noun, but I did so here as it's capitalized elsewhere.  The most important thing is to be consistent across the whole paper.}
%
  \item \textit{Interpretable assessment of the generative model:}  We explored the generative characteristics of the model by Monte-Carlo simulation and forecasting, which helps understand the causal physical process that describes the thermal behavior of a house. We also tested the quality of the models by forecasting indoor temperature with the learned building parameters. 
 % \item \textit{Effect of prior selection:} We studied their behavior with rigorous experiments beliefs about a building's condition in the form of prior selection to further enrich the model. 
  \item \textit{Effects of transfer learning \& prior selection:} 
 We proposed a transfer learning based approach by learning buildings' parameters in summer, when HVAC is typically operational in cooling mode, and used it to aid fitting the data in seasons when HVAC is not used or operates in heating mode. We propose a systematic approach to prior selection to incorporate beliefs about the buildings' characteristics in the model and conducted rigorous experiments to study their behavior. 
    \end{itemize}
The rest of the paper is structured as follows. In Section~\ref{sec:Related} we discuss related work on building parameter identification and Bayesian estimation. In Section~\ref{sec:Model} we propose the Bayesian State Space model for building parameter identification. In Section~\ref{sec:Analysis} we present two case studies and provide analysis of the model and finally conclude in Section~\ref{sec:Conclusion}.

\section{{Related Works}}   
\label{sec:Related}
  In this section, we review the previous works in three major related areas -- parametric modeling of buildings' thermal dynamics, techniques for parameter estimation, and a brief review of techniques for Bayesian inference.  \\
\indent \textbf{Parametric modeling of buildings' thermal dynamics}  helps to understand a  building's quality with few parameters. There are three approaches for modeling buildings' thermal dynamics -- \emph{White-box}, \emph{Black-box } and \emph{Gray-box} modeling. \emph{White-box} approach models all physical processes of a building~\cite{HONG2000347,KIRCHER2015454} by formulating exact system dynamics. Such deterministic models are difficult to construct as the exact dynamics are often unavailable and due to the presence of noise in the data, arising from unaccounted factors. \emph{Black-box modeling} approaches, such as regression, neural networks etc., are applied to model indoor temperature as a function of observed data, much like outdoor temperature~\cite{nielsen2010analysis}. However they do not describe the generative process and thus isn't effective for interpretation. A \emph{gray-box model} is  a combination of prior physical knowledge and statistical approaches. The heat dynamics of the building was formulated  using several equivalent models of varying complexity in~\cite{bacher2011identifying}, that  estimated the insulation and the thermal mass of a building. An extension of such an approach included the effect of wind speed on infiltration is proposed in~\cite{EstimationBuildsys}, and expansionary effect of air with temperature changes was modeled in~\cite{siemann2013performance}.    \\ 
\indent  \textbf{ Parameter estimation} for the gray-box model was performed in~\cite{KRISTENSEN2004225} by  maximum likelihood estimation (MLE) and maximum a posteriori estimation (MAP). An extension of the approach~\cite{ghosh2015modeling}, chose a simpler model to represent the dynamics and learned the residuals separately with Gaussian priors. Other works have focused on optimization based techniques~\cite{Optimization,EstimationBuildsys}, where the objective is to minimize  deviation between measurements and predictions from the model. Although, these approaches are simpler, they do not incorporate noise estimation in the equations. Alternatively, a Bayesian approach offers a natural way of dealing with parameter uncertainty in a state space model~\cite{gordon1993novel, geweke2001bayesian}. Bayesian methods have been widely used for the closely related problem of building energy modeling ~\cite{LI2016194,heo2012calibration,kim2013stochastic,kristensen2017bayesian}, but have been less well studied in the context of \emph{thermal modeling for buildings}~\cite{bacher2011identifying,rouchier2018calibration}.  A majority of previous  works have applied the Metropolis-Hastings algorithm for Bayesian inference ~\cite{rouchier2018calibration,heo2012calibration,kim2013stochastic,LI2016194,kristensen2017bayesian} which is ill-fitted for the specific problem as it takes a large number of steps to achieve convergence. The No U-Turn sampler (NUTS) showed better results~\cite{chong2017bayesian} for parameter estimation in a related problem, building energy models, so we choose the latter.\\
\indent  \textbf{Bayesian Inference}, as performed in the previous works, used uninformed uniform priors and/or Normal priors for the model parameters~\cite{bacher2011identifying,rouchier2018calibration}. Such assumptions do not hold true as the parameters typically have non-Normal distributions. Non-normal priors do not have conjugacy with the  likelihood, and analytical solutions of the posterior distribution are not possible. In such cases, algorithms that do not rely on conjugacy become important such as MCMC and Variational Inference. MCMC algorithms are capable of overcoming this problem but are time-consuming. Alternatively, Variational Inference~\cite{beal2003variational, luttinen2013fast} is an approximate inference that derives a lower bound for the marginal likelihood which can be optimized using stochastic gradient descent. In our experiments, we use the Mean Field Variational Inference and find that it provides similar parameter estimation to MCMC algorithms.

\section{{Proposed Approach}}\label{sec:Model}  
\begin{figure}[h]
\includegraphics[scale=.20, width=.45\textwidth]{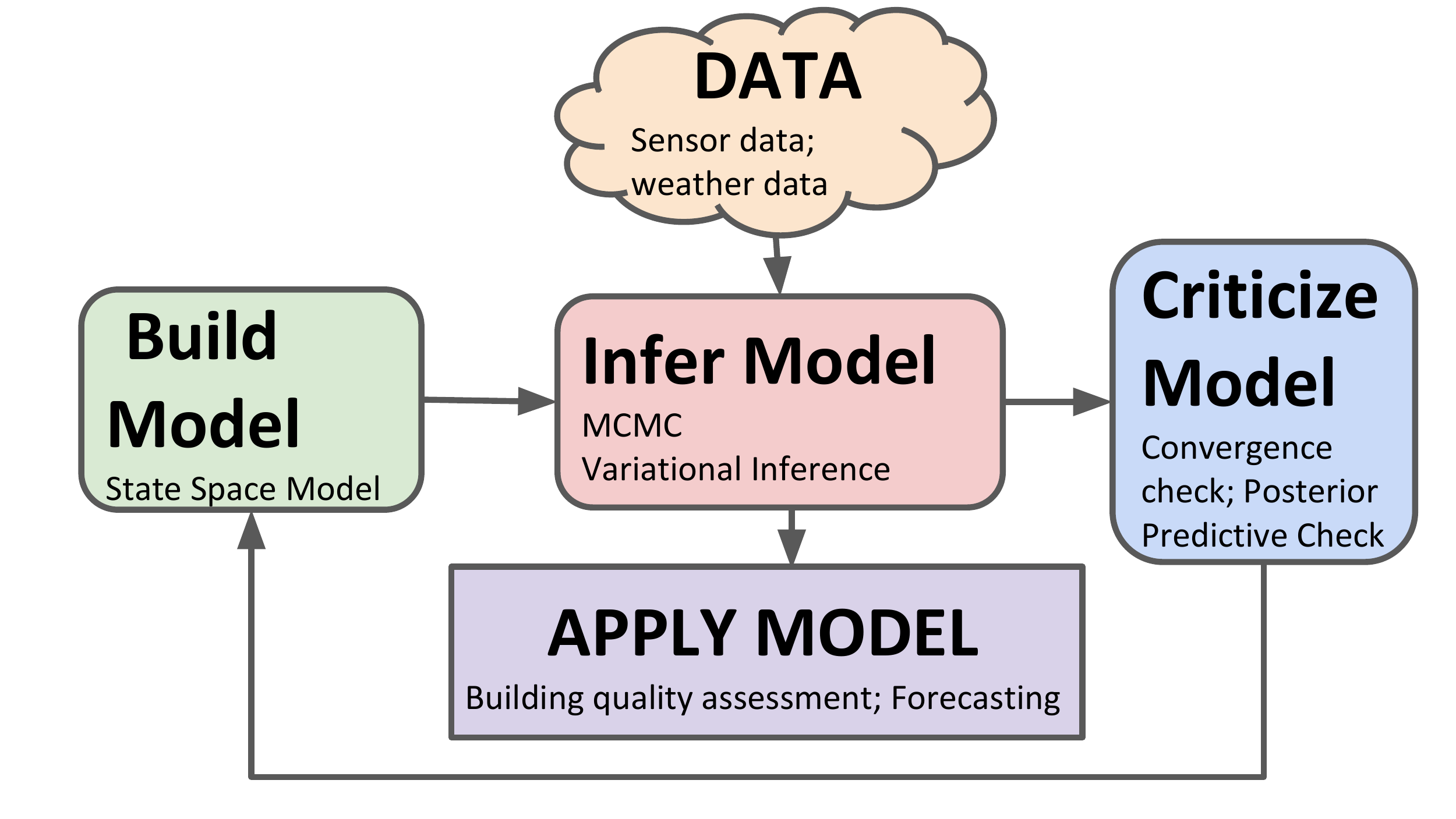}
\caption{Box's Loop}
\label{fig:Box}
\end{figure}

We follow the iterative modeling approach known as \emph{Box's loop}~\cite{blei2014build}, shown in Fig.~\ref{fig:Box}, for estimating buildings' thermal parameters. The process starts with a collected and pre-processed dataset.  We propose a Bayesian state space model to frame the problem and estimate parameters using MCMC and Variational Inference. Finally, we test for model convergence and measure the goodness of fit.
  
\subsection{{ Bayesian Linear State Space Model} } 

  The generalized linear state space models consist of a sequence of $M$-dimensional observations ($y_1$, $y_2$, ... $y_N$), assumed to be generated from latent $D$-dimensional states $X$ = ($x_1$, $x_2$, ... $x_N$) and control variables $U$ =  ($u_1$, $u_2$, ... $u_N$). The data $Y$ is generated by the following state space equations:
 
   \begin{eqnarray}
   x_n = A x_{n-1} + B u_{n-1} + \mathcal{N}(0,Q) \label{eqn:stateEvolution} \\
   y_n = C x_n +  \mathcal{N}(0,R) \mbox{ ,} \label{eqn:obs}
    \end{eqnarray}
    
where Eqn.~\ref{eqn:stateEvolution} is the state evolution equation (analogous to HMM state transition) and Eqn.~\ref{eqn:obs} is the observation or measurement equation (emission probability). The overall state transition probability is given as

 \begin{eqnarray}
  P(X | A, B) = \mathcal{N}(x_0 | m_0, \Lambda_0^{-1}) \times  \nonumber \\
  \prod_{i=1}^N \mathcal{N}(x_n | Ax_{n-1} + B u_{n}, diag(\tau_1^{-1}))\mbox{ ,} \label{eqn:Transition}
 \end{eqnarray}

    where $x_0$ is an auxiliary initial state with mean $m_0$ and a precision matrix of $\Lambda_0$ (the matrix inverse of the covariance matrix). The emission probability is given by  
  
    \begin{equation}
    P(Y | C, X , \tau) = \prod_{n=1}^N \mathcal{N}(y_n | C x_n, diag(\tau^{-1})) \mbox{ .}\label{eqn:Emission}
    \end{equation}

\begin{figure}[H]
  \includegraphics[scale=.25,width=.45\textwidth]{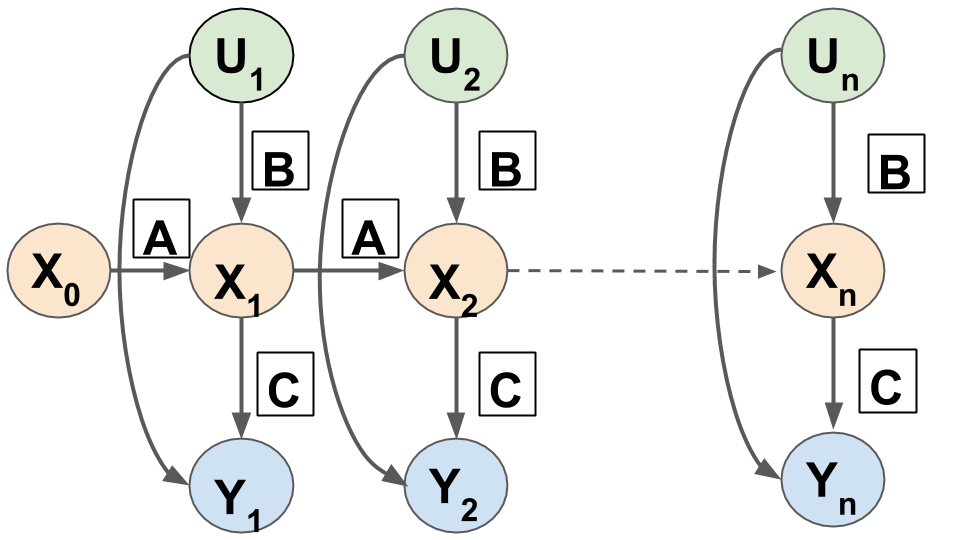}
  \caption{Bayesian State Space Model}
  \label{fig:BSSM} 
  \end{figure}

   Here $Y$ is a normal distribution with mean\textbf{ CX }and a covariance matrix with $diag(\tau^{-1})$. The covariance matrix \textbf{R} is a diagonal matrix as the noise is independent of the observed states $Y$. 
The graphical nature of the BSSM model is shown in Fig~\ref{fig:BSSM}, which is analogous to an Input-Output HMM~\cite{bengio1995input}.
    
  \subsection{{Problem Formulation}}  \label{subsec:Problem} 
\begin{figure}[h]
\includegraphics[width=.45\textwidth]{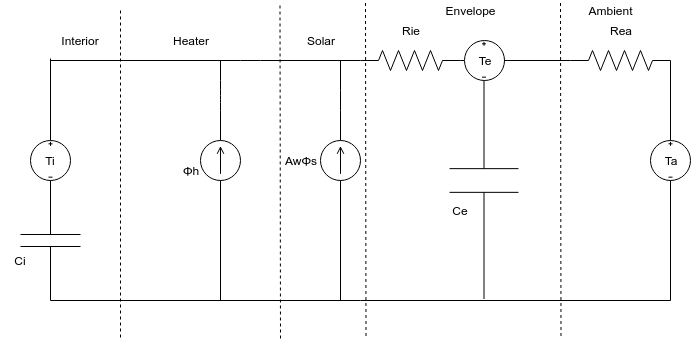}
\caption{TiTe circuit model}
\label{fig:TiTe}
\end{figure}

We use an example to illustrate how to formulate a building's thermal equations and incorporate them into the proposed state space model framework. Figure~\ref{fig:TiTe} shows an equivalent circuit that describes the thermal dynamics of a house.

  In this example, called the $TiTe$ model, we assume that there are two latent state spaces $Ti$ and $Te$ that describes the indoor and envelope temperatures. The thermal dynamics is represented by a set of stochastic differential equations derived from the equivalent assumption. The equations of the process are given by: 
  
   \begin{flalign}
   dT_i &= \frac{1}{R_{ie}C_i} (T_e - T_i) dt + \frac{1}{C_i}\Phi_hdt + \frac{A_w }{C_i}\Phi_s dt +  \sigma_i d\omega_i \\
   dT_e &= \frac{1}{R_{ie}C_e} (T_i - T_e) dt +  \frac{1}{R_{Bea}C_e} (T_a - T_e) dt   +  \sigma_e  d\omega_e \\
   Y_k &= T_{i_k} + e_k\mbox{ ,}
   \end{flalign}

 where \textit{t} is the time, \textit{$R_{ie}$} is the thermal resistance between the interior and the building envelope, \textit{$R_{ea}$} is the thermal resistance between the building envelope and the ambient air, $C_i$ is the heat capacity of the interior, $C_e$ is the heat capacity of the building envelope, $\Phi_h$ is the energy flux from the heating system, $A_w$ is the effective window area, $\Phi_s$ is the energy flux from solar radiation, $T_a$ is the ambient air temperature, $\{\omega_{i,t}\}$ and $\{\omega_{e,t}\}$ are standard Wiener processes with variances $\sigma_i$ and $\sigma_e$  respectively, where $t$ is the point in time of a measurement. $Y_t$ is the indoor temperature, $T_{i_k}$ is the measured interior and $e_k$ is the measurement error, which is assumed to be a Gaussian white noise process. Converting the differential equations (Eqns 5--7) as difference equations we get the transition and emission matrix form as:\\

  \begin{flalign}
   \begin{bmatrix} Ti(t+1) \\ Te(t+1)\end{bmatrix} &=
\begin{bmatrix}
    1 -  \frac{1}{R_{ie}C_i}    &  \frac{1}{R_{ie}C_i}   \\
   \frac{1}{R_{ie}C_e}     &    1 -  \frac{1}{R_{ie}C_e} -  \frac{1}{R_{ia}C_e}  
\end{bmatrix} \times \begin{bmatrix} Ti(t) \\ Te(t)\end{bmatrix}  \nonumber \\ 
&+ \begin{bmatrix}
    0   &  \frac{1}{C_i} &   \frac{A_{w}}{C_i} \\
   \frac{1}{R_{ia}C_e}     &    0 & 0
\end{bmatrix}\times \begin{bmatrix} Ta(t+1) \\ \Phi_h(t+1) \\ \Phi_s(t+1) \end{bmatrix}  + \begin{bmatrix}
   \sigma_{i}    &  0   \\
   0   &    \sigma_{e}  
\end{bmatrix} \label{eqn:trans}  \\
     Y(t)  &=  \begin{bmatrix} 1 & 0\end{bmatrix} \times \begin{bmatrix} Ti(t) \\ Te(t)\end{bmatrix} + \sigma \mbox{ .} \label{eqn:emis}
  \end{flalign}  
  \normalsize

   Eqn~\ref{eqn:trans} is the state transition of the dynamic system and is equivalent to the general form as presented in Eqn~\ref{eqn:stateEvolution} that gives us the transition probability, i.e $P(x_{t+1} | x_t)$ $\sim$ $\mathcal{N}(AX + BU, \Sigma_w)$ (Eqn~\ref{eqn:Transition}). Similarly,  Eqn~\ref{eqn:emis} is equivalent to the measurement equation provided by Eqn~\ref{eqn:obs} that gives us the emission probability (Eqn~\ref{eqn:Emission}).  In Eqn~\ref{eqn:trans}, the first two matrices models the physical dynamics and the third matrix is the measure of stochasticity in the data. Similarly, in Eqn~\ref{eqn:emis}, the first matrix is the measurement equation and $\sigma$ is the error in measurement. In the base case, we assume an uninformative Gamma prior over the model parameters  and the hyper-parameters of the gamma distribution are automatic relevance determination (ARD) parameters, which prune out components that are not significant enough. We provide broad priors to the gamma distribution by setting the shape and rate to a very small value~\cite{beal2003variational}. Thus parameters are given as   $\mathbf{R} \sim \Gamma(\alpha = \delta, \beta = \delta)$,
    $\mathbf{C} \sim \Gamma(\alpha = \delta, \beta = \delta)$, 
    $\mathbf{Aw} \sim \Gamma(\alpha = \delta, \beta = \delta)$,  where $\delta$ is a very small value. We also impose a bound on the parameters, which can help by limiting the parameters to certain reasonable ranges. We formulate other instantiations of the physical models in the case studies presented in Section 4.1.2. 
 \subsection{{Bayesian Inference}}
 \label{subsec:Inference}
 \indent   Bayesian inference recovers the posterior distribution over parameters and latent variables of the model, which can hence be used to perform prediction. While exact solutions can be achieved for some basic models, computing the posterior distribution is generally an intractable problem, in which case approximate inference is needed.\\
\indent \textbf{Markov chain Monte Carlo (MCMC)} algorithms are a widely applied method for approximate inference, which aims to estimate the posterior using a collection of samples drawn from an appropriate Markov chain.  Hamiltonian Monte Carlo (HMC)~\cite{neal2011mcmc} algorithms such as NUTS avoid the random walk behavior by taking a series of steps informed by first-order gradient information. These features allow it to converge to high-dimensional target distributions much more quickly than simpler methods such as random walk Metropolis Hastings~\cite{yildirim2012bayesian}.  The No U-Turn Sampler (NUTS)~\cite{hoffman2014no} uses a recursive algorithm to build a set of likely candidate points that span a range of the target distribution, stopping automatically when it starts to backtrack and retrace its steps, which prevents the revisiting of previously explored paths. In this work, we select the NUTS sampler for inference. \\  
\indent Another option is \textbf{Variational Inference}, which is a class of algorithms that are deterministic alternatives to MCMC. This reduces inference tasks to an optimization problem~\cite{blei2017variational}. In a probabilistic latent model setting, Y is the observed data, X is the latent variable space and $\theta$ the model parameters. An approximating distribution $q(X,\theta)$ over the latent variables and parameters, called the \emph{variational distribution}, is constructed to approximate the posterior. The objective is to reduce the ``gap'' between the variational and the posterior distribution. This gap is given by the Kullback-Leibler divergence, which is the relative entropy between the two distributions, given as:

    \begin{align} \label{eqn:KL}
      KL (q(X,\theta) || p(X, \theta | Y)) \nonumber = \mathbb{E}_q  \Big \lbrack log \: \frac{q(X,\theta)}{p(X, \theta | Y)} \Big \rbrack    \nonumber \\
       = \mathbb{E}_q \lbrack log\: q(X,\theta)\rbrack -  \mathbb{E}_q \lbrack log\: p(X, \theta, Y) \rbrack + log\: p(Y) \mbox{ .}  \\\nonumber
    \end{align}
   \normalsize
    
   In Eqn~\ref{eqn:KL}, $log\: p(Y)$ is independent of the distribution $q(X, \theta )$, so minimizing Eqn~\ref{eqn:KL} is equivalent to maximizing: 
   
\begin{align} \label{eqn:ELBO}
     \mathbf{L}(q)  = \mathbb{E}_q  \lbrack log\: p(X, \theta, Y) \rbrack -\mathbb{E}_q \lbrack log\: q(X,\theta)\rbrack  \nonumber \\
     = \mathbb{E}_q  \lbrack log\: p(X, \theta, Y) \rbrack + H(q) \mbox{ .} \\\nonumber
\end{align}
 \normalsize

\indent Using Jensen's inequality, $\mathbf{L}(q)$ can be shown to be a lower bound on $\log\: p(Y)$, and is hence known as the Evidence Lower Bound (ELBO). To make inference tractable, we make simplifying assumptions on $q$.  The most commonly used assumption is the mean-field approximation, which assumes that the latent variables are independent of each other. Thus the variational distribution with $N$ latent variables is assumed factorized as $q(X,\theta)   =  q(\theta)\prod_{i=1}^Nq(X_i) $. Traditionally, a Variational Inference algorithm requires developing and implementing model specific optimization routines. Automatic Differentiation Variational Inference (ADVI)~\cite{kucukelbir2017automatic} proposes an automatic solution to posterior inference. ADVI first transforms the model into one with unconstrained real-valued latent variables. It then recasts the gradient of the variational objective function as an expectation over \textbf{q}.  This involves the gradient of the log of the joint likelihood with respect to the latent variable $\bigtriangledown_{\theta}\;log\;p(x,\theta)$, which is computed using reverse-mode automatic differentiation~\cite{maclaurin2016modeling}. This gradient term is applied to optimize the parameters using a stochastic gradient descent approach.  \\
\indent It is important to note the underlying assumptions of ADVI. It factors the posterior distribution such that all the state variables are statistically independent, following the mean-field approximation. For a highly correlated posterior, e.g. in state space models, where the intuition is that  $x_{t+1} \sim N(x_{t}, \theta)$ will be highly correlated with $x_t$, the mean-field assumption is rather unrealistic. The method can still work well in practice, however, as the (uncorrelated) $q$ is fit to the (correlated) $p$, thereby exploiting dependencies, even though they are not ultimately encoded in $q$. NUTS, on the other hand, is very good at exploring a correlated, high-dimensional distribution, but can suffer in both run-time and convergence speed versus ADVI. We empirically evaluate the effectiveness of these approximations by comparing the parameters inferred by both the methods.

  \subsection{{Model Criticism }}
  
Model criticism requires tests for convergence   and testing goodness of fit on held out data. Since the primary objective of the study is to obtain the estimated parameter values, we also inspect the credible interval of the parameters. If the region is too wide we infer that the uncertainty in estimation is high.  \\ 
\indent \textbf{Convergence Diagnostics:} We select the \textit{Gelman-Rubin diagnostic}~\cite{gelman2011inference}, which checks for the lack of convergence by comparing the variance between multiple chains to the variance within each chain. Convergence is more straightforward to analyze for Variational Inference. The convergence criterion is simply to iterate until the ELBO no longer increases. \\ 
\indent \textbf{Goodness of fit} is tested using \emph{posterior predictive checks}, which are performed by simulating replicated data under the fitted model and then comparing these to the observed data to look for systematic discrepancies between real and simulated data~\cite{gelman2013bayesian}.\\ 
\indent \textbf{ Credible Interval:} The motivation behind using a Bayesian approach is to find the range of possible values for the building parameters. A standard measure of confidence in some (scalar) quantity $\theta$ is the ``width'' of its posterior distribution. This can be measured using a 100(1 - $\alpha$)\% \emph{credible interval}, where we select $\alpha$ as 0.05 to estimate parameters with a 95\% probability,
   \begin{equation}    
   C_{\alpha}(D)=(	l , u) : P(	 l \leq \theta \leq u| Y) = 1 -  \alpha \mbox{ ,}
   \end{equation}
    where the interval for a parameter is bounded by (l,u) with a probability $1-\alpha$. The credible interval is a Bayesian alternative to a frequentist confidence interval. A frequentist keeps the parameters fixed and varies the confidence interval whereas a Bayesian approach is to keep the credible region fixed and vary the model parameters.

   \subsection{{Application of the Models}}

    \subsubsection{{Exploration}}

    In terms of building modeling, we are primarily interested in learning the different \textbf{R} and \textbf{C} parameters. As we consider different multi-state lumped models, the cardinality of the sets \textbf{R} and \textbf{C} may vary but the overall values should remain the same. To find the composite resistance of the equivalent circuit, the resistance and capacitance are obtained by Kirchoff's law. However, the simple addition or geometric sum required to compute the composite parameters cannot straightforwardly be done as Bayesian Inference provides random variables rather than scalar quantities. The distribution of the sum of two random variables can be obtained by the  convolution of their density functions.

\begin{algorithm}
\caption{Indoor Temperature Forecast}
\label{alg:Forecast}
\begin{algorithmic}[1]
\Procedure{Forecast\text{ }} {\emph{Input}: Distribution of the model parameters $\theta=\{\mathbf{R} , \mathbf{C}, \mathbf{A}, \mathbf{X}_{T}\}$, time window of forecast $K$}
\State Set start temperature state to $X_T$ 
\State  \{$R_i$ , $C_i$, $A_i$\} <- Draw N Sample(R, C, A) 
\For{i in 1 : N}
\State   $X^{(i)}_{T+1:T+K}$ $\leftarrow$ $BSSM(X_T, R_i , C_i, A_i)$
\EndFor
\State Mean Prediction $\leftarrow$ $\mathbf{E}\{ X^{(i)}_{T+1:T+K} \}$
\State Credible Interval $\leftarrow$ \{$Max(X^{(i)}_{T+1:T+K})$ , $Min(X^{(i)}_{T+1:T+K})$\}
\EndProcedure
\end{algorithmic}
\end{algorithm}

\subsubsection{Forecasting}
      
  We perform 24 hour ahead forecast, after learning the parameters of the model. We assume that a outdoor temperature forecast data is given to us and we assumed that the HVAC is operational in the last known mode. The forecasting and prediction of HVAC time is given in Algorithm~\ref{alg:Forecast}.    When the HVAC is set to a particular temperature and assuming that it is not changed within the horizon of the forecast, then the indoor temperature will be centered around the set-point in a range known as the thermostat hysteresis setting. In general, the range is $\pm$ $\delta$ lies within 0.5 -- 1 $^\circ$F. We sample from the estimated parameters' distributions to obtain the forecasting interval.

\subsection{Implementation}
   
   We implemented the Bayesian State Space model using the PyMC3 probabilistic programming library in Python~\cite{PYMC3}. PyMC3 is built on  Theano~\cite{Theano} and has built-in implementations for MCMC algorithms and Variational Inference methods. We formulated the different components of the state space model and set the prior distributions for the model parameters. We deployed our methods on a system with 16 GB RAM system and I7 processor. The initial version of the codes is available in the \href{https://github.com/Nilavro/BSSPy}{\color{blue} BSSP} Github repository.

\begin{table*}[t]
\centering
\caption{Results of Study I}
\label{Table:Study1}
\begin{tabular}{@{}p{0.6cm}p{0.8cm}p{0.6cm}p{0.6cm}p{0.6cm}p{0.6cm}p{0.6cm}p{0.6cm}p{0.6cm}p{0.6cm}p{0.6cm}p{0.6cm}p{0.6cm}p{0.8cm}p{0.7cm}p{2.5cm}@{}}
\hline
Model                   & Method & Total R     & Total C                                                   & Ria         & Rie                                                      & Rih                                                     & Rea                                                    & Ci                                                       & Ce                                                       & Ch                                                        & Aw                                                     & Ae                                                    & NRMSE (\%) & RMSE        & Convergence Time (Steps) \\ \hline
\multirow{3}{*}{Ti}     & EKF     & 5.29        & 24.797                                                      & 5.29        &                     -                                     &      -                                                   &                                  -                      & 2.06                                                     &                                 -                         &       -                                                    & 7.89                                                    & -                                                      & 0.4        & 0.06        & 2.68s ( Steps = 39)               \\
                        & MCMC   & 5.29, 0.06  & 24.96,  0.36   & 5.29, 0.06  &                -                                          &                                      -                   &     -                                                   & 24.96, 0.36                      -                        &      -                                                    &                                               -            & 7.95, 0.675                                            &                                          -             & 0.8     & 0.10   &   30 min (Chains = 4, Steps = 5000)      \\
                        & ADVI   & 5.29, 0.67  & 24.75, 0.40    & 5.29, 0.067 &      -                                                    &                              -                           &   -                                                     & 24.73, 0.43                                    &  -                                                        &                                       -                    & 7.87, 0.675                                            &                                  -                     & 0.4       & 0.05       & 2.26min (Steps = 180000)         \\ \hline
\multirow{3}{*}{TiTe}   & EKF     & 5.36       & 39.17                                                      &      -       & 5.17                                                   &                               -                          &  0.19                                                    & 19.99                                                     & 19.18                                                  &       -                                                    & 23.78                                                  &   -                                                    & 0.3        & 0.04        & 14.65s  (100)            \\
                        & MCMC   & 5.27, 0.17  & 89.95,  7.92   &        -     & 1.73, 0.056                                              &                          -                               & 3.54, 0.122                                            & 21.39, 0.30                                              & 68.56, 7.91                                              &                 -                                          & 10.75, 0.64                                            &            -                                           & 0.3      & 0.04      &  1.5 hrs (4,5000)      \\
                        & ADVI   & 5.29, 0.002 & 25.31, 0.86   &       -      & 1.98, 0.001                                              &                               -                          & 3.31, 0.002                                            & 24.49 , 0.502                                            & 0.82, 0.008                                              &                      -                                     & 7.90, 0.86                                             &                  -                                     & 0.3      & 0.04       & 2:38 min (Steps =  180000)         \\ \hline
\multirow{3}{*}{TiTeTh} & EKF     &        -  & -                                                     &           -  &                                          -            &                  -                               &                              -                      &                  -                                   &                                      -               &                       -                         &    -                                               &    -                                                &    -    &  -    & No Convergence              \\\\
                        & MCMC   &       -      &             -                                              &          -   & 159.13, 273.83 & 70.23, 159.81 & 23.08, 68.44 & 119.77, 206.30 &  121.85, 184.87  &  313.33,   342.47  &  15.52,  28.71  &  0.092,  0.60 &     -       &    -         & No Convergence           \\
                        & ADVI   & 2.8, 0.004  &  184.45,   13.58 &     -        & 2.176, 0.010                                             & 0.23, 0.003                                             & 0.63, 0.003                                            & 177.82, 83.22                                            & 2.05, 0.03                                               & 4.59, 0.097                                               & 52.76, 193.86                                          & 45.50, 181.90                                         &   -     &      -   & 4.23min  (Steps = 180000)        \\ \hline
\end{tabular}
\end{table*}

  \section{{Analysis}} 
  \label{sec:Analysis}
  In this section we provide two case studies. In the first test case we compare the results with small dataset to contrast and compare the gray-box models' solutions with the Kalman filter and Bayesian state space model. In the second case study, we present results and analyses on larger scale data from the Dataport Dataset~\cite{Pecan}.
  
  \subsection{{Case Study I: Exploratory Study on a Benchmark Dataset}}
    \label{subsec:Case}
   
\subsubsection{Dataset} We compare the results with the benchmark dataset provided   in~\cite{bacher2011identifying} and the circuit assumptions of the house mentioned in the paper. The data is from a Flexhouse in Ris{\o} DTU in Denmark, and  was collected during a series of experiments carried out in February to April 2009, where measurements consist of five minute values over a period of six days. The dataset consist of a single signal representing the indoor temperature ($y$ $^{\circ}$C).  Observed ambient air temperature at the climate station ($Ta$ $^{\circ}$C).  Total heat input from the electrical heaters in the building ($\Phi_h$ kW). The global irradiance was measured at the climate station ($\Phi_s$ kW/$m^2$).

 \subsubsection{Problem Formulation}
 \label{subsec:ProbForm}
 
 First we constructed all the models suggested in~\cite{bacher2011identifying}. The  CTSM~\cite{kristensen2003continuous}  package can be used to model Continuous Time Stochastic Processes   which is realized using an Extended Kalman Filter (EKF).   We define \textbf{R}, \textbf{C}, and \textbf{A} to be the set of resistances, capacitances and area of solar infiltration for individual models. The three models which we chose for inspection and their system dynamics as follows: 
 \begin{itemize}
     \item \textit{Ti Model:} Here the house as a whole is assumed to have one thermal resistance ($R_{ia}$) and capacitance ($C_i$).
 
 \begin{flalign}
 T_i(k+1)  &=  (1 - \frac{1}{R_{ia}C_i}) \times T_i(k)  + \frac{1}{R_{ia}C_i} \times T_e(k+1)  \nonumber    \\ &+ \frac{1}{C_i} \times \Phi_h(k+1) + \frac{A_w }{C_i}\Phi_s(k+1) +  \sigma_i  \\
 Y_k &= T_i(k)  + e_k 
 \end{flalign}
 
 \item \textit{TiTe Model:}    We provided a detailed description of formulation using the TiTe model in Section~\ref{subsec:Problem}. 
 
 \item \textit{ TiTeTh Model:} The three state model represents the interior subscripted by i, the exterior subscripted by e  and the heater subscripted by h. The formulation for the three states are as follows:

  \begin{flalign}
   T_i(k+1)  &=  (1 - \frac{1}{R_{ie}C_i}) \times T_i(k)  + \frac{1}{R_{ie}C_i}\times  T_e(k+1) \nonumber   \\  &+\frac{A_w }{C_i} \times  \Phi_s(k+1) + \sigma_i    \\
   T_e(k+1) &= (1-\frac{1}{R_{ie}C_e}-\frac{1}{R_{ea}C_e}) \times  T_e(k)  +  \frac{1}{R_{ea}C_e} \times  T_a(k+1) \nonumber   \\ &+ \frac{1}{R_{ie}C_e} \times  T_i(k+1) +  \sigma_e    \\
     T_h(k+1)  &=  (1 - \frac{1}{R_{ih}C_h}) \times  T_h(k) + \frac{1}{R_{ih}C_h} \times  T_i(k+1) \nonumber   \\  &+ \frac{1}{C_h} \times  \Phi_h(k+1) + \sigma_h   \\
   Y_k &= T_i(k)  + e_k 
   \end{flalign} 
 
 \end{itemize}

 \subsubsection{{Results Discussion}} We show the results 
 of the first case study in Table~\ref{Table:Study1}. We provide the \textbf{mean} and \textbf{variances} ($\mu$, $\sigma$) of the estimated model parameters --  \textbf{R}, \textbf{C} and \textbf{Aw}, within a 95\% credible interval range as shown in Table~\ref{Table:Study1}. The total \textbf{R} and total \textbf{C} the composite thermal resistance and capacitance of the building. We compare the results of Bayesian Inference with the point estimates with an Extended Kalman Filter (EKF). The insights from the study are as follows: \\
\indent \textbf{{\emph{Estimated model parameters:}}} From the Table~\ref{Table:Study1} we found that the mean of the credible interval for the estimated parameters for the Bayesian approaches is similar to that of the EKF point estimate. The EKF assumes the parameters to have uniform priors and thus performs MLE for estimation. An approximate ADVI provides a similar parameter estimation as that of an equivalent run of MCMC inference. \\
\indent \textbf{{\emph{Comparison with the point estimates:}}} A direct comparison of the model's performances between the Bayesian methods and the EKF is difficult. We take the mean of the parameter estimated from Bayesian inference and then perform a one-step-ahead prediction and compare that with the EKF. The metrics used for comparison are the root mean squared (RMSE) and the normalized root mean squared errors (NRMSE) of the one-step-ahead prediction and we find that estimates from ADVI give us the best results (Table~\ref{Table:Study1}). \\
\indent \textbf{{\emph{Time of execution to reach convergence:}}} MLE estimates of the EKF is the fastest as it does not require computation of the full posterior distribution, however, it does not provide estimation error over model parameters.  The MCMC algorithm is the most time consuming one, where we increase the number of steps and check for convergence using Gelman-Rubin diagnostic. We selected 4 chains and an initial burn in 5000 steps which is intended to give the Markov Chain time to reach its equilibrium distribution when there is a random initial starting point.  Compared to MCMC, ADVI is much faster. For the $TiTeTh$ model, we obtain no convergence for the EKF or MCMC, i.e. the credible intervals are very wide. But ADVI provides reasonable intervals for some parameters. We listed the time of execution for the different approaches in Table~\ref{Table:Study1}.  

 \begin{figure}[t] 
\begin{subfigure}{.35\textwidth}
\includegraphics[width=\textwidth]{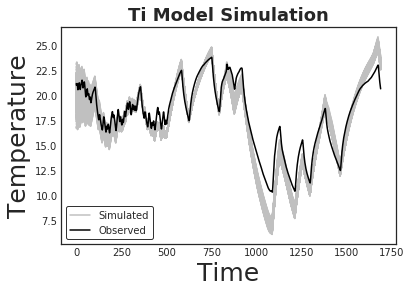}
\label{fig:RollingADVITi}
\caption{Ti model}
\end{subfigure}
\hfill
\begin{subfigure}{.35\textwidth}
\includegraphics[width=\textwidth]{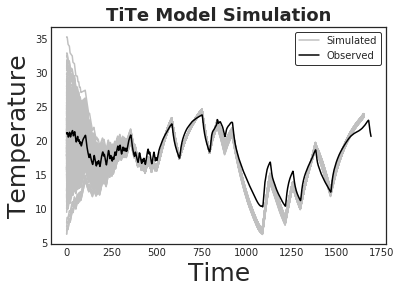}
\label{fig:RollingADVITiTe}
\caption{TiTe model}
\end{subfigure}
\caption{Monte Carlo Simulation of Indoor Temperature}
\label{fig:Sim}
\end{figure}

 \indent\textbf{{\emph{Monte-Carlo simulation:}}} We perform a Monte-Carlo simulation to generate the possible indoor temperature scenarios when weather and HVAC usage is provided. In Figure~\ref{fig:Sim} we show the results of the simulated prediction for the Ti and TiTe models, drawing samples from the inferred parameter distributions. We considered the starting state to be drawn from a $\mathbb{N}$(70, 5) distribution, i.e. our guess for the indoor temperature will be within the range of 60 - 80$^{\circ}$F. The simulated prediction shows that the actual value of the indoor temperature is enclosed within the credible region. It, however, deviates in certain sections, which we hypothesize is because the thermal mass of a house can change with varying temperature. The RC constant~\cite{boylestad2002electronic}  of the data changes with time as the thermal mass \textbf{C} of a house can vary, due expansion (or contraction) of air. A more generalized formulation of thermal dynamics will require exploring longitudinal studies that to correlate between the parameter and temperature changes with the heater and cooler usage for long duration. 
  \begin{figure}[t] 
\begin{subfigure}{.45\textwidth}
\includegraphics[width=\textwidth]{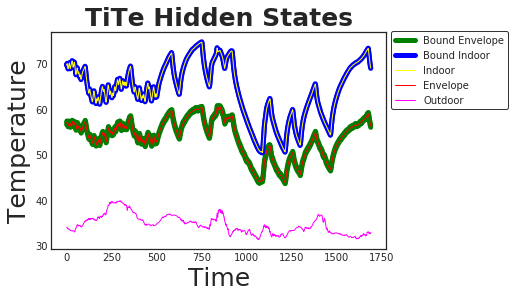}
\caption{Hidden States of TiTe  model}
\label{fig:HiddenTiTe}
\end{subfigure}
\hfill 
\begin{subfigure}{.5\textwidth}
\includegraphics[width=\textwidth]{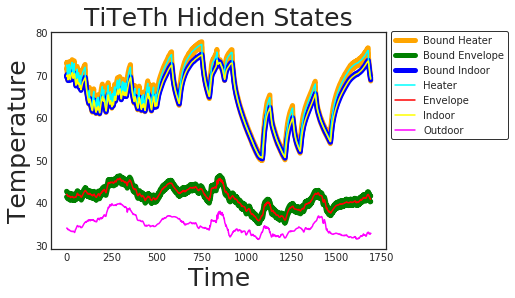}
\caption{Hidden States of TiTeTh model}
\label{fig:HiddenTiTeTh}
\end{subfigure}
\caption{Visualizing Hidden State Dynamics}
\label{fig:Qual}
\end{figure}

 \begin{figure*}[t]
\centering     
\begin{subfigure}{.38\textwidth}
\includegraphics[width=\textwidth]{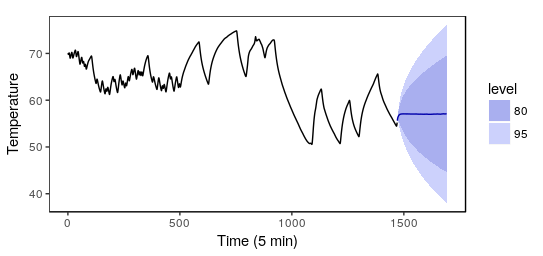}
\caption{ARIMAX}
\label{fig:ARIMAX}
\end{subfigure}
\begin{subfigure}{.3\textwidth}
\includegraphics[width=\textwidth]{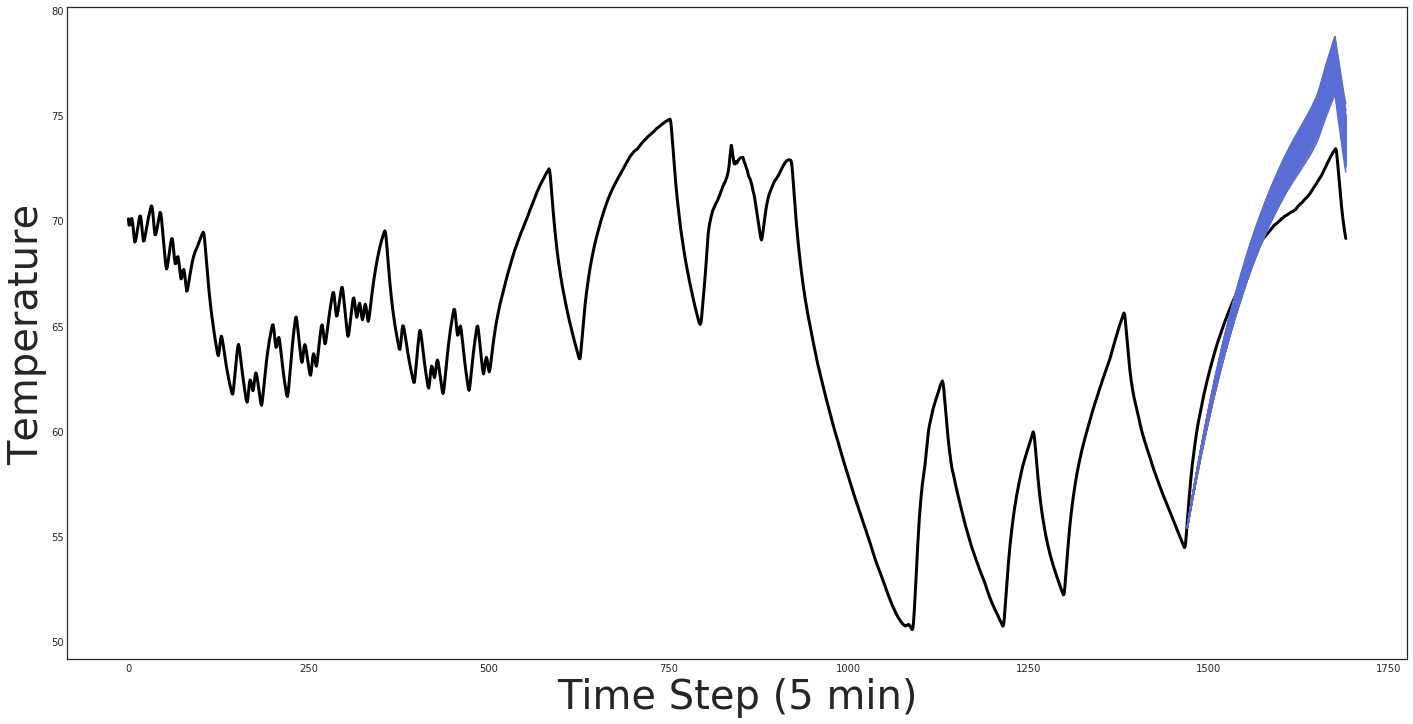}
\caption{Ti Model}
\label{fig:TiForecast}
\end{subfigure}
\begin{subfigure}{.3\textwidth}
\includegraphics[width=\textwidth]{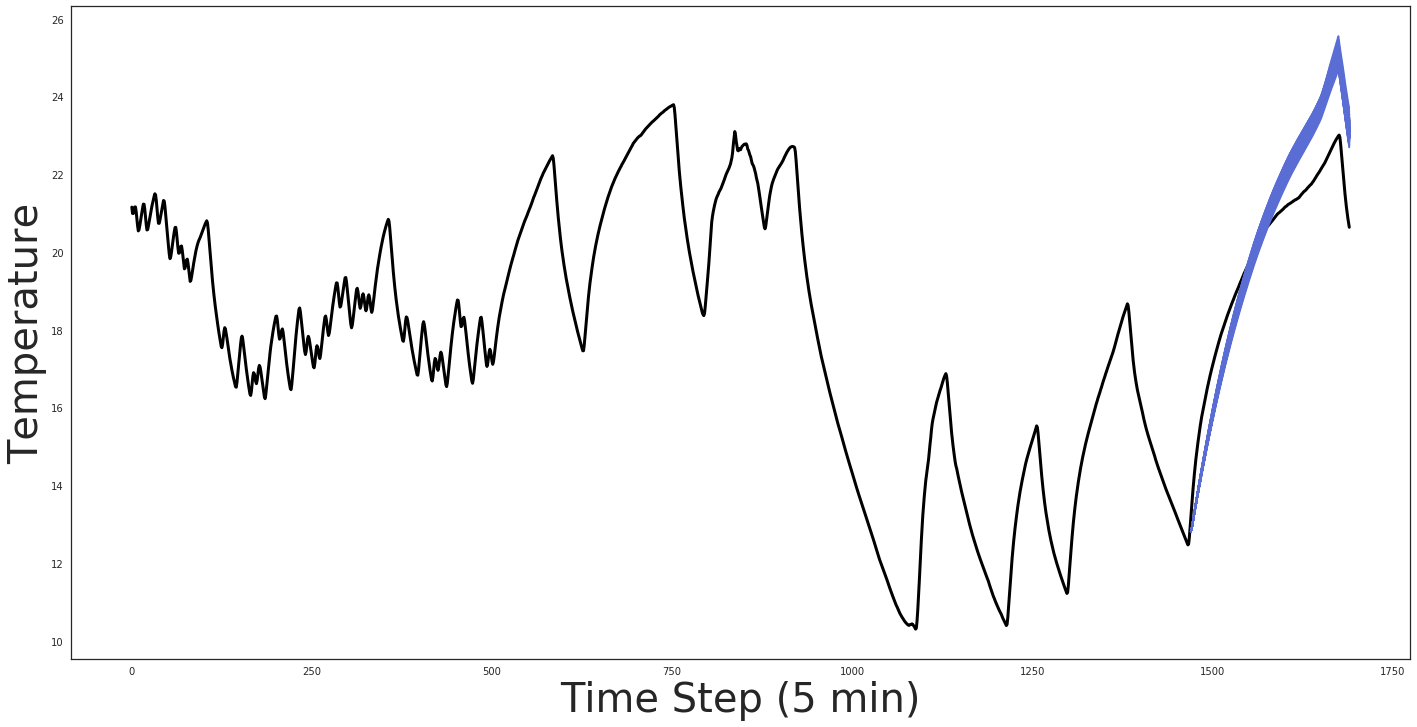}
\caption{TiTe Model}
\label{fig:TiTeForecast}
\end{subfigure} 
\caption{Comparison of Forecasting}
\end{figure*}

\indent  \textbf{{\emph{Qualitative assessment of the hidden states:} }}In Figure~\ref{fig:Qual}, we show the learned hidden states of the two-state $TiTe$ and three state $TiTeTh$ model. For the $TiTe$ model shown in Fig~\ref{fig:HiddenTiTe}, show that the estimated hidden state for the envelope is sandwiched between the indoor temperature and the outdoor temperature and is more correlated with the indoor temperature. Whereas in the $TiTeTh$  model (Fig~\ref{fig:HiddenTiTeTh}) the envelope state is more correlated with the outdoor temperature. However, the heater's temperature is the same as that of the indoor temperature, which implies it does not capture an independent factor of the  hidden state space. The plots in Fig~\ref{fig:Qual} show the error margin of the hidden states  obtained from the highest posterior distribution.

 \indent \textbf{{\emph{Forecasting:}}}   We compared the forecasting results of the BSSM with an auto-regressive integrated moving average with exogenous variables (ARIMAX). We used the first 5 days for training the BSSM and learn the building parameters. We then used the learned parameters to obtain a day ahead forecast within 95\% prediction interval, as presented in Algorithm~\ref{alg:Forecast}.  Our assumption is that the heater stays in the same state as the last known state and assumed that the solar radiation and temperature data are available.  In Figures~\ref{fig:ARIMAX} --~\ref{fig:TiTeForecast} we show the output of forecasting for the ARIMAX, $Ti$ and $TiTe$ models. For quantitative evaluation we chose the mean absolute percentage error (MAPE) to find the error in mean of the forecast and calculated the percent of data within 95\% forecast interval as shown in Table~\ref{table:Forecast}. The mean forecast error is lower in case of the BSSM models as they better learn the dynamics of the process. However, as the model parameters have a narrow credible interval, the actual data lies outside the forecast interval in  but provides a narrower band for which the actual value partially lies outside the credible interval. In contrast, the forecasting result of ARIMAX has less correlation, although the actual forecast is within the confidence interval. 
 
\begin{table}
\centering
\caption{Results of Forecasting}
\begin{tabular}{|l|l|l|}
\hline
Method &  MAPE  & \% of data within 95\%\\ &&forecast interval \\ \hline
ARIMAX & 0.19 & 100\%\\ \hline
Ti Model & 0.05 & 78\% \\ \hline
TiTe Model & 0.04 & 79\%\\ \hline 
\end{tabular}
\label{table:Forecast}
\end{table}

\subsection{{Case Study II: Prior Selection \& Transfer Learning}} 

\label{subsec:Case2}

\subsubsection{{Dataset}} The Dataport dataset is a publicly available dataset, created by Pecan Street Inc, which contains building-level electricity data from 1000+ households. We performed our experiments on three single-family homes from Texas (\emph{dataid} = 484, 739, 1507) based on metadata availability and proper registration of indoor temperature and HVAC usage data. The metadata, which has information about 52 homes, provides a general understanding about the buildings and helps us create prior distributions over the R-values.  Here, House 739 does not have heating data available and the metadata does not include a measure for House 1507's R-value.  
\subsubsection{{Experimental setup}} In this case study we explore the effects of prior selection and transfer learning. The two processes are inherently tied together, since in the Bayesian approach, ``today's posterior becomes tomorrow's prior.'' Our approach here is to learn the parameters from the AC usage season, where data is more consistent, and transfer the learned parameters as priors to seasons when HVAC has typically no usage and/or operates in heating mode. We investigate the effect of three sets of priors:

\begin{itemize}
\item \textit{Informed Priors (Set 1):} We selected informed gamma priors. This is useful when we have some notion about the parameters' values such as an initial audit to estimate the R-value of a building. We select a strong prior on the R-value where the mean of the R-value is same as that of the estimate and the standard deviation is 1.  

\item \textit{Hyper Priors (Set 2):} In this set, we don't have a direct estimation regarding the buildings' parameters but have a vague understanding about the expected value from the metadata. We encode such beliefs by setting  a hyper-prior for the mean, that is sampled selected from a mixture of lognormal distributions. We empirically found that R-values are a mixture of lognormal distributions, conditioned on the year built and conditioned square foot, by performing a maximum likelihood estimate. The estimated parameters of the two lognormal distributions as shown in Fig~\ref{fig:hist}, are ($\mu_0$, $\sigma_0$) = (3.02, 0.59) and ($\mu_1$, $\sigma_1$) = (3.43, 0.50),  respectively. 

\item \textit{Uninformed Priors (Set 3):} Finally, in \textit{Set 3}, we chose \textit{uninformed} flat gamma priors for the R-values, where, we have no knowledge of the buildings' parameters.  In all three cases we set a flat gamma prior on the C-values.
\end{itemize}

 \begin{table*}[h]
\centering
\caption{Results of Case Study II}
\begin{tabular}{@{}lllllllll@{}}
\toprule
Homes & HVAC Mode & R-Value & \multicolumn{2}{l}{Informed Priors} & \multicolumn{2}{l}{Hyper Priors} & \multicolumn{2}{l}{Uninformed Priors} \\ \midrule
 &  &  & R & C & R & C & R & C \\ \cmidrule(l){4-9} 
 & No Usage &  & 26, 0.02 & 85.92, 12 & 53.24, 14.99 & 81.6, 10.86 & 449.04, 431.90 & 80.82 , 10.71 \\
484 & AC & 26 & 26, 0.02 & 85.03 , 12.5 & 52.12, 15.33 & 80.81, 11.08 & 453.73, 428.96 & 80.25, 10.99 \\
 & Heater &  & 26, 0.02 & 132.44 , 12.79 & 75.87, 15.9 & 89.35, 11.7 & 343.63 , 276.16 & 81.48 , 11.73 \\
 & No Usage &  & 6.1, 0.23 & 73.15, 18.84 & 61.82, 14.46 & 15.43, 1.945 & 62.1 , 14.33 & 15.33, 2.005 \\
739 & AC & 6 & 6.1, 0.23 & 64.09, 20.7 & 58.88, 10.87 & 15.55, 1.79 & 59.47, 10.43 & 15.65, 1.84 \\
 & Heater &  & - & - & - & - & - & - \\
 & No Usage &  & - & - & 24.52, 0.215 & 103.4, 7.34 & 48.66, 20.91 & 107.42, 7.56 \\
1507 & AC & NA & - & - & 23.58, 0.215 & 103.52, 7.89 & 47.65, 21.95 & 107.88, 7.55 \\
 & Heater &  & - & - & 23.58, 0.23 & 104.65, 8.79 & 52.22, 26.62 & 108.84, 8.38 \\ \bottomrule
\end{tabular}
\label{table:BestParam}
\end{table*}

\indent For all three cases, we set an upper bound on the R-values to be 70, which we found from the metadata. We assign an uninformed gamma prior on the C values. For all cases, we initially estimate for the AC usage scenario and use the mean and variance of the estimated parameters to set the prior for the other seasons.  The sign of the heat flux, as provided in Eqn~\ref{eqn:trans}, is negative when AC is used and the HVAC is in cooling mode. We do not have an exact value for heater's flux but we use the furnace which provides the binary ``ON-OFF'' signal of the heater and multiply an extra unknown parameter $\Phi_h$ to estimate the heat flux. Similarly to the previous section, we estimate all parameters within a 95\% credible interval. We also varied the size of the dataset of sizes [200, 500, 1000, 2000, 5000].
\subsubsection{Results}    
\begin{figure*}[!htp] 
\begin{subfigure}{.48\textwidth}
\includegraphics[width=\textwidth]{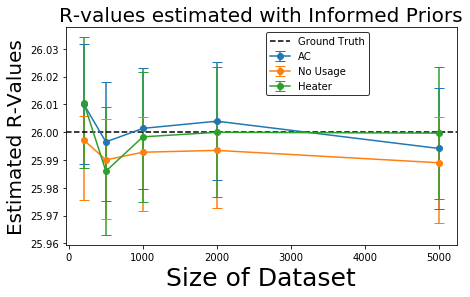}
\caption{}
\end{subfigure}
\hfill 
\begin{subfigure}{.48\textwidth}
\includegraphics[width=\textwidth]{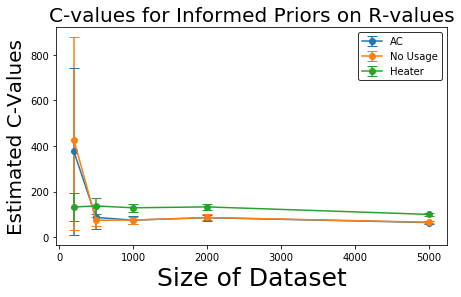}
\caption{}
\end{subfigure}
\caption{Informed Prior set on R-values}
\label{fig:ErrIn}
\end{figure*}

\begin{figure*}[!htp]
\begin{subfigure}{.48\textwidth}
\includegraphics[width=\textwidth]{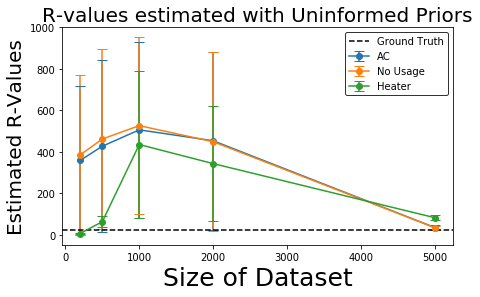}
\caption{}
\end{subfigure}
\hfill 
\begin{subfigure}{.48\textwidth}
\includegraphics[width=\textwidth]{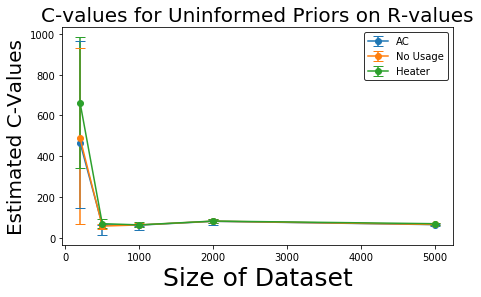}
\caption{}
\end{subfigure}
\caption{Uninformed set on R-values}
\label{fig:ErrUn}
\caption*{ \textmd{{These figures show the results of parameter estimation of R and C values for House 484 with varying data sizes and different prior selection.}}}
\end{figure*}

The prior selection directly influences the value of the parameters, parameter transfer and depends on the size of the dataset. A summary of the results is presented in Table~\ref{table:BestParam} for 2000 data points, which provides us the most likely parameter estimates. We present the result in the form of mean and the error margins i.e. ($\mu$, $\pm \epsilon$). We find that the Informed Priors provide us with the most consistent estimates both across size of the datasets and when we perform transfer learning. As shown in Fig~\ref{fig:ErrIn}, the informed parameters remain consistent with the change in the size of the dataset with very little margin of error ($\pm$ 1). Parameter transfer also works best when informed priors are applied (Fig~\ref{fig:ErrIn}), but can provide us different estimates when being transferred from AC usage to Heater usage seasons.  The hyper-priors reduce the margin of error when applied for smaller datasets. For example, in Fig~\ref{fig:ErrUn}, the R-values have large error margins when uninformed priors are chosen, which is significantly reduced when hyperpriors were used Fig~\ref{fig:ErrHp}.  

\subsubsection{Discussion}

\begin{figure*}[!htp]
\begin{subfigure}{.48\textwidth}
\includegraphics[width=\textwidth]{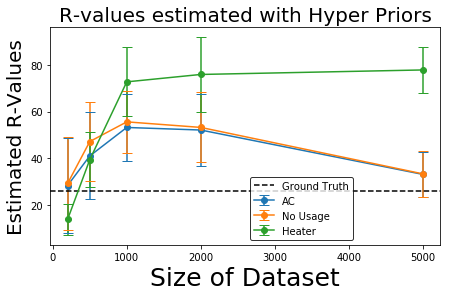}
\caption{}
\end{subfigure}
\hfill 
\begin{subfigure}{.48\textwidth}
\includegraphics[width=\textwidth]{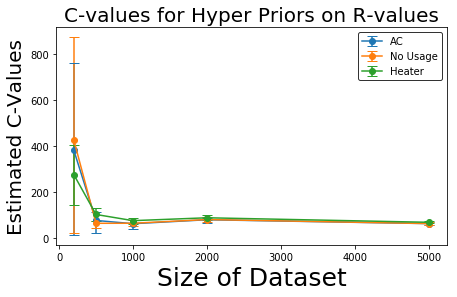}
\caption{}
\end{subfigure}
\caption{Hyper Prior set on R-values}
\label{fig:ErrHp}
\end{figure*}

Information in the data overwhelms prior information not only when the size of the dataset is large, but also when the prior encodes relatively small information. For example in House 739 (Table~\ref{table:BestParam}), an approach using uninformed priors will try to get the best estimate that fits the data, but the parameters may not be accurate. For this case, a sharp prior centered around an initial estimate gives the best result. Uninformative priors are easily persuaded by data, while strongly informative ones may be more resistant. When the size of the dataset is small, hyperpriors effectively reduce the margin of error in parameter estimation Fig.~\ref{fig:ErrHp}.
 
% \jf{Convergence time in MCMC iterations, or in size of the dataset?  I don't think we have results showing the former, and it wouldn't match my intuition.  Assuming you mean the latter, can put a supporting reference to Figure 8 here.}
 
\subsection{General Recommendations}%Takeaways from the Studies} % \jf{From both studies, plural, right?} 

     Based on our studies, we recommend constructing a Bayesian state space model customized for the problem at hand, carefully selecting the system dynamics and  priors. We suggest using ADVI for parameter estimation as it provides similar estimates but is faster than MCMC. In realistic settings, it is better to perform an initial audit to determine the home's insulation parameters and fix an informed prior on the parameter set. We suggest to use informative priors, if enough metadata is available to set them reliably. However, if the objective is to monitor a large set of homes, we recommend setting a hyper prior based on the beliefs from a sample of the dataset. If the heat flux is known from the HVAC, learning from one season and applying it to another can improve estimation. %safely\jf{too strong a word?} learn from one season and apply it on another. 

\section{Conclusion \& Future Work}
\label{sec:Conclusion}
 
 In this paper, we proposed and  systematically studied Bayesian statistical approaches to buildings' thermal parameter estimation. We developed a generalized state-space modeling framework that integrates building physics equations with a statistical model. The model estimates buildings' structural parameters which influence the indoor temperature conditioned on HVAC usage and weather factors. We contrast model learning using MCMC and ADVI algorithms and show that Variational Inference is faster and provides a similar estimation to MCMC. A visual inspection of the hidden states was employed to assess the model dynamics, and we found that merely increasing model complexity does not capture any significant factors of the thermal characteristics. We further showed the model's applications, such as simulating probable outcomes and forecasting the future. The effects of prior selection on the parameter estimation were studied in detail. We found that informed priors provide the best estimates, but when such information is not present prior beliefs can help to better learn the models. Also, we found that priors are key to transfer learning, and model parameters learned from one season can be used to model thermal dynamics under the condition that %exogenous data is available and are properly scaled
 properly scaled exogenous data is available. \\ 
   \indent The focus of our future research is in two directions.  We are presently instrumenting several homes with smart thermostats and temperature sensors. This study serves as a guide to large-scale analysis as we attempt to further incorporate air leakages and construct room level thermal behavior analysis. We plan to learn from the data that is being collected longitudinally and incorporate the learned models in NEST thermostats to monitor homes' condition continuously. Secondly,  we will focus on incorporating air-leakage into the framework and  correlating with standardized metrics such as $ACH_{50}$. Common air-infiltration models (e.g. LBL model~\cite{sherman1992superposition}), have complex non-linear characteristics for which we will explore non-linear state space models.

\bibliographystyle{ACM-Reference-Format}
\bibliography{Main}

\end{document}